\documentclass{article}
\usepackage[utf8]{inputenc}
\usepackage{geometry}
\usepackage{graphicx}
\usepackage{booktabs}
\usepackage{natbib}
\usepackage{amsmath}
\usepackage{amsfonts}
\usepackage{array}
\usepackage{float}
\usepackage{xcolor}
\usepackage{caption}
\usepackage{subcaption}
\usepackage{url}
\usepackage[hidelinks]{hyperref}
\definecolor{H_blue}{HTML}{1F77B4}
\definecolor{M_orange}{HTML}{FF7F0E}
\definecolor{C_green}{HTML}{2CA02C}
\definecolor{myteal}{HTML}{327273}

\title{ST-GRU PredRNN-V2: A GRU-Based Reformulation of PredRNN-V2 for Video Prediction}
\author{Augusta van Haren \\ {\normalsize\itshape Radboud University} \and Hanna Hoogen \\ {\normalsize\itshape Radboud University} \and Luca Pattavina \\ {\normalsize\itshape Radboud University} \and Quincy Rommen \\ {\normalsize\itshape Radboud University}}
\date{October 2026 \\[6pt] \normalsize\itshape{All authors contributed equally.}}

\begin{document}
\maketitle

\begin{abstract}
\noindent Video prediction requires models to capture complex spatiotemporal dependencies and to retain relevant information over long sequences. PredRNN and its successor PredRNN-V2 address this challenge with spatiotemporal Long Short-Term Memory (ST-LSTM) units that maintain two separate memory states, at the cost of a relatively complex recurrent building block.

In this work, we investigate whether the simpler gating of Gated Recurrent Units (GRUs) can be transferred to spatiotemporal predictive learning. To this end, we develop a spatiotemporal GRU (ST-GRU) unit as a GRU-based alternative to the ST-LSTM. The ST-GRU replaces the ST-LSTM's three gates per memory state with two GRU-style gates (update and reset), and merges the temporal memory state into the hidden state, which reduces the number of parameters by about 5\%. We incorporate the ST-GRU into PredRNN-V2 and evaluate the resulting model on the Moving MNIST video prediction benchmark, following the original experimental protocol.

Averaged over ten predicted frames, the ST-GRU-based PredRNN-V2 achieves an MSE of 55.1, an SSIM of 0.883, an LPIPS of 0.065 and a PSNR of 19.53\,dB. It does not surpass the original ST-LSTM-based PredRNN-V2 on MSE, SSIM or PSNR, but it achieves a lower LPIPS and performs on par with the earlier ST-LSTM-based PredRNN, while using a simpler recurrent cell. These results suggest that ST-GRUs are a promising building block for spatiotemporal predictive learning and motivate further work on their design and integration. 
\end{abstract}

\section{Introduction}
Recent developments in Artificial Intelligence (AI) have enabled the synthetic creation of images with remarkable success \citep{betker2023improving}. Generating video frames, however, remains challenging, due to complex spatiotemporal dependencies within the data and uncertainty in predicting future motion \citep{Cai_2018_ECCV}. This task, known as predictive learning, focuses on generating future data points conditioned on data points from the past. For video, the aim is to predict upcoming frames based on a sequence of past frames.

Modelling sequential data was a key motivation for developing recurrent neural networks (RNNs), which capture temporal dependencies by maintaining a hidden state over time. However, RNNs suffer from a fundamental bottleneck: the vanishing gradient problem limits their ability to retain long-term dependencies, particularly in long input sequences \citep{hochreiter1998vanishing}. As a consequence, the network forgets information that is relevant for its future predictions, which reduces their accuracy.

Two gated architectures were developed to alleviate this problem: Long Short-Term Memory (LSTM) units and Gated Recurrent Units (GRUs). The LSTM adds a memory cell controlled by three gates (an input gate, a forget gate and an output gate), which together decide what information to store, discard and expose. This allows the network to retain temporal dependencies over much longer time spans \citep{hochreiter1997long}. The GRU is a later development that follows the same principle of selective information retention, but simplifies the architecture: it has only an update gate and a reset gate, and no separate memory cell \citep{chung2014}.

LSTMs and GRUs have been widely used for modelling sequential data such as language and speech. Video prediction, however, additionally requires models to preserve spatial structure, representing both low-level appearance and high-level dynamics in order to generate future frames. Convolutional variants such as ConvLSTM \citep{shi2015-convlstm} and ConvGRU \citep{ballas2015-convgru} address this by replacing the fully connected transformations inside the recurrent units with convolutions.

Building on this, \citet{wang2017-predrnn} developed the Predictive Recurrent Neural Network (PredRNN), which was later extended to PredRNN-V2 \citep{wang2021-predrnnv2}. Instead of classical LSTMs, PredRNN is built on spatiotemporal LSTM (ST-LSTM) units that maintain not one but \textit{two} memory states. The way these ST-LSTM units are chained allows spatiotemporal information to flow both across time steps and across the layers of the network. Through this mechanism, top-down expectations and bottom-up signals are combined to predict future inputs. This differs from stacked ConvLSTMs, in which memory states are confined to individual recurrent units. PredRNN(-V2) outperformed previous architectures on several video prediction benchmarks.

Although PredRNN(-V2) predicts spatiotemporal dynamics well across various datasets, its ST-LSTM units are relatively complex, with three gates per memory state and two separate memory states. GRUs, in contrast, use fewer gates than LSTMs while typically achieving comparable performance on sequential tasks \citep{chung2014}. This work explores whether this simplification also transfers to spatiotemporal predictive learning.

To this end, we develop a GRU-based alternative to the ST-LSTM, which we call the spatiotemporal GRU (ST-GRU). We replace PredRNN-V2's ST-LSTM units with ST-GRUs and compare the resulting model with the original PredRNN-V2, as well as with several other recurrent baselines, on video frame prediction on the Moving MNIST dataset. Concurrently with our work, \citet{dong2025-stgru} proposed a different ST-GRU cell for the original PredRNN; we include their results in our comparison and discuss how the two approaches differ. Our code is available at \url{https://github.com/xPatta/ST-GRU}.

\section{Methods}

\subsection{ST-LSTM-based PredRNN}
PredRNN \citep{wang2017-predrnn} consists of a stack of ST-LSTMs with a dual-memory mechanism that learns both the spatial structure and the temporal dynamics of sequential data, allowing it to predict future frames.

The ST-LSTMs in PredRNN handle four flows of information: the input state $\mathcal{X}_t$, the hidden state $\mathcal{H}_t^l$, and \emph{two} memory states, $\mathcal{C}_t^l$ and $\mathcal{M}_t^l$. These memory states are decoupled. The temporal memory $\mathcal{C}_t^l$ operates on a slower timescale and is designed to learn long-term dependencies in the data. It is passed horizontally from time step $t-1$ to time step $t$ within the same layer $l$. The spatiotemporal memory $\mathcal{M}_t^l$, in contrast, adapts more rapidly to changes in the input and captures short-term dynamics. It follows a deeper transition path that zigzags upwards, passing information from layer $l-1$ to layer $l$ within the same time step; only the top layer at time step $t-1$ is connected to the bottom layer at time step $t$.

Each memory state has its own input gate $i_t$ ($i_t'$), forget gate $f_t$ ($f_t'$) and input modulation gate $g_t$ ($g_t'$). The input gate, with a sigmoid activation ($\sigma$), determines how much new information from the current input is added to the memory. The forget gate decides what information is discarded. The input modulation gate generates candidate values using the hyperbolic tangent ($\tanh$); these are scaled by the input gate as $i_t\odot g_t$ and added to the memory state, replacing the forgotten information.

Finally, the updated memory states $\mathcal{C}_t^l$ and $\mathcal{M}_t^l$ produce the new hidden state $\mathcal{H}_t^l$ in two ways: indirectly via the output gate $o_t$, and directly via concatenation, a $1\times1$ convolution and a $\tanh$. The ST-LSTM equations are given in Figure~\ref{fig:ma_mathematicalcomparison_ST-LSTM}, and schematics of the cell and its integration in PredRNN(-V2) are shown in Figures~\ref{fig:ma_visualcomparison11_ST-LSTM} and~\ref{fig:ma_visualcomparison22_ST-LSTM} (Appendix~\ref{app:ma}).

\subsection{ST-LSTM-based PredRNN-V2}
\label{ST-LSTM-based PredRNN-V2}
PredRNN-V2 \citep{wang2021-predrnnv2} extends PredRNN in three ways. First, a decoupling loss explicitly increases the distance between the updates of the two memory states in latent space, encouraging them to learn different aspects of the spatiotemporal variations and preventing redundant features. Second, reverse scheduled sampling is a curriculum learning strategy for the encoder that gradually shifts its input from the model's own predictions to the ground-truth frames, which forces the model to learn long-term dynamics. Third, PredRNN-V2 supports action-conditioned video prediction.

\subsection{ST-GRU-based PredRNN-V2}
\label{ST-GRU-based-PredRNN-V2}
We propose a variant of PredRNN-V2 in which the ST-LSTM building blocks are replaced by GRU-based counterparts, which we call ST-GRUs. We design the ST-GRU around the key characteristics of a GRU, while keeping the separation between temporal and spatiotemporal memory flows of the ST-LSTM \citep{wang2017-predrnn,wang2021-predrnnv2}.

Following the GRU, each memory update in the ST-GRU uses only two gates: an update gate $z_t$ (respectively $z'_t$) and a reset gate $r_t$ (respectively $r'_t$). In addition, the ST-GRU omits the separate temporal memory state $\mathcal{C}_t^l$ and integrates its function into the hidden state $\mathcal{H}_t^l$, since both follow the same horizontal path from one time step to the next.

As a result, the two pairs of gates act on $\mathcal{H}_t^l$ and $\mathcal{M}_t^l$, rather than on $\mathcal{C}_t^l$ and $\mathcal{M}_t^l$ as in the ST-LSTM. The update gate $z_t$ decides how much of the state is replaced by new content, whereas the reset gate $r_t$ decides how much of the previous state is used when computing that new content. Together they play roles similar to the input and forget gates of the ST-LSTM \citep{chung2014}. The gates of both states share the convolutions applied to $\mathcal{M}_t^{l-1}$, which further reduces the number of parameters. The gates are used to compute the candidate states $\widetilde{\mathcal{H}}_t^l$ and $\widetilde{\mathcal{M}}_t^l$, which then produce the new hidden state $\mathcal{H}_t^l$ indirectly via the output gate $o_t$, and directly via concatenation, a $1\times1$ convolution and a $\tanh$. The ST-GRU equations are given in Figure~\ref{fig:ma_mathematicalcomparison_ST-GRU}, and schematics of the cell and its integration in PredRNN(-V2) are shown in Figures~\ref{fig:ma_visualcomparison11_ST-GRU} and~\ref{fig:ma_visualcomparison22_ST-GRU} (Appendix~\ref{app:ma}), next to their ST-LSTM counterparts. The figures highlight both what the two cells share (the decoupled spatiotemporal memory) and where they differ (the number of gates and memory states).

\subsection{Implementation}
We implemented the ST-GRU-based PredRNN-V2 on top of the official PyTorch implementation of PredRNN-V2 \citep{github-predrnn}. Following the structure of the original ST-LSTM cell, we implemented the ST-GRU as designed in Section~\ref{ST-GRU-based-PredRNN-V2}, with two instead of three gates per memory state and without using $\mathcal{C}_t^l$. The variable for $\mathcal{C}_t^l$ is retained in the code only to keep the interface compatible with PredRNN-V2. We then built a version of PredRNN-V2 that uses the ST-GRU as its building block, passing information through ST-GRUs instead of ST-LSTMs. Because the ST-GRU has no temporal memory update to decouple from $\mathcal{M}_t^l$, we trained this model without PredRNN-V2's decoupling loss, using only the MSE loss; reverse scheduled sampling was kept unchanged. All other settings follow the official Moving MNIST configuration: four recurrent layers with 128 hidden channels each, $5\times5$ convolutions, a patch size of 4, a batch size of 8 and a learning rate of $10^{-4}$. With these settings, the ST-GRU-based PredRNN-V2 has 22.6M parameters, compared with 23.9M for the ST-LSTM-based PredRNN-V2, a reduction of about 5\%. Our implementation is available at \url{https://github.com/xPatta/ST-GRU}.

\subsection{Experimental Setup}
\label{experimentalsetup}
To compare the ST-GRU-based PredRNN-V2 directly with the original ST-LSTM-based version, we repeated the Moving MNIST experiment of \citet{wang2021-predrnnv2}, following the same training and testing protocol. Specifically, we trained the model for 80,000 iterations using the official training configuration \citep{github-predrnn}, and evaluated it with four metrics: mean squared error (MSE), structural similarity index measure (SSIM) \citep{wang2004-ssim}, learned perceptual image patch similarity (LPIPS) \citep{zhang2018-lpips}, and peak signal-to-noise ratio (PSNR).

We further compared our model with several other recurrent models. ConvLSTM \citep{shi2015-convlstm} and ConvGRU \citep{ballas2015-convgru} are convolutional LSTM- and GRU-based RNNs \emph{without} decoupled spatiotemporal memory. The ST-LSTM-based PredRNN \citep{wang2017-predrnn} has decoupled spatiotemporal memory but \emph{without} the V2 additions (see Section~\ref{ST-LSTM-based PredRNN-V2}). Finally, \citet{dong2025-stgru} concurrently proposed an ST-GRU cell and integrated it into PredRNN. Because they evaluated it with the same experimental protocol as above, we include their reported results in our comparison. We emphasise that their ST-GRU-based PredRNN differs from our ST-GRU-based PredRNN-V2 \emph{both} in the design of the ST-GRU cell and in the absence of the V2 additions (see Section~\ref{ST-LSTM-based PredRNN-V2}). Results for the LSTM-based baselines are taken from \citet{wang2021-predrnnv2}, and those for the GRU-based baselines from \citet{dong2025-stgru}. PSNR is not reported in these sources, so we computed it for the ST-LSTM-based PredRNN-V2 using the official trained model \citep{github-predrnn}.

\section{Results}

\begin{figure}[htb]
    \centering
    \includegraphics[width=\textwidth]{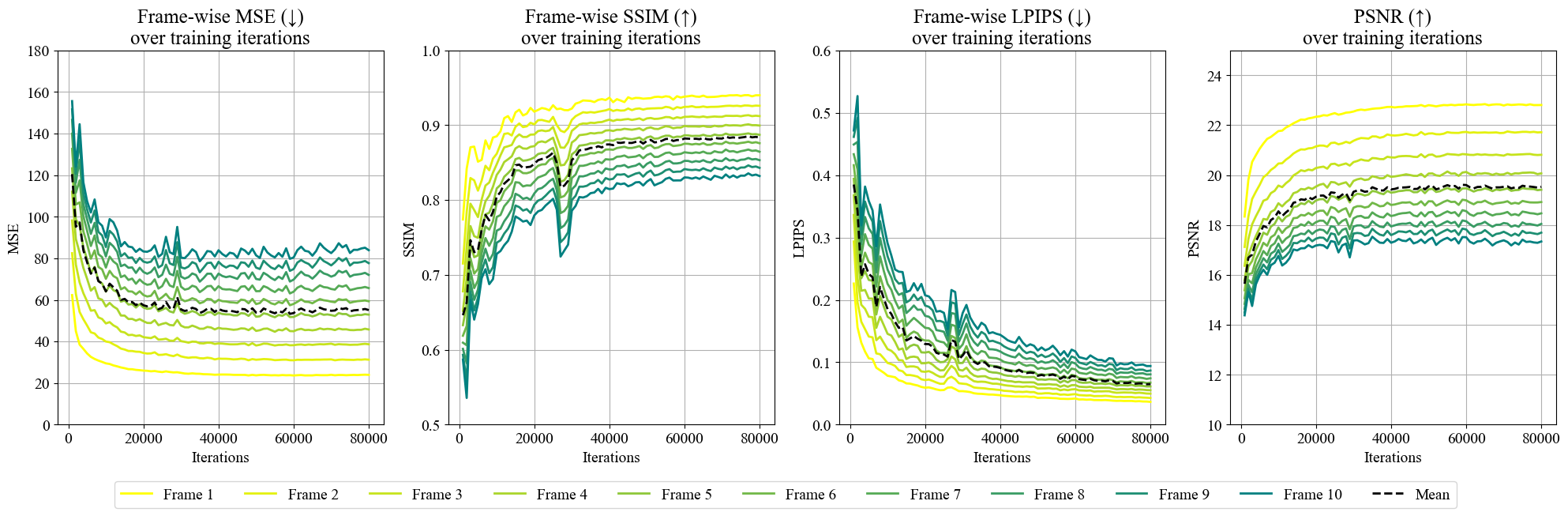}
    \caption{Frame-wise metrics (MSE, SSIM, LPIPS, PSNR) of the ST-GRU-based PredRNN-V2 on the Moving MNIST test set over 80,000 training iterations. The frame number indicates the number of time steps between the predicted frame and the last observed frame.}
    \label{fig:results_st-gru_training}
\end{figure}

As shown in Figure~\ref{fig:results_st-gru_training}, all four frame-wise metrics (MSE, SSIM, LPIPS and PSNR) plateau within the 80,000 training iterations, indicating that the number of iterations was sufficient for convergence. Early frames are consistently predicted more accurately than later frames. This is expected, as early frames are temporally closer to the input sequence and are therefore inherently easier to predict.

\begin{figure}[H]
    \centering
    \includegraphics[width=\textwidth]{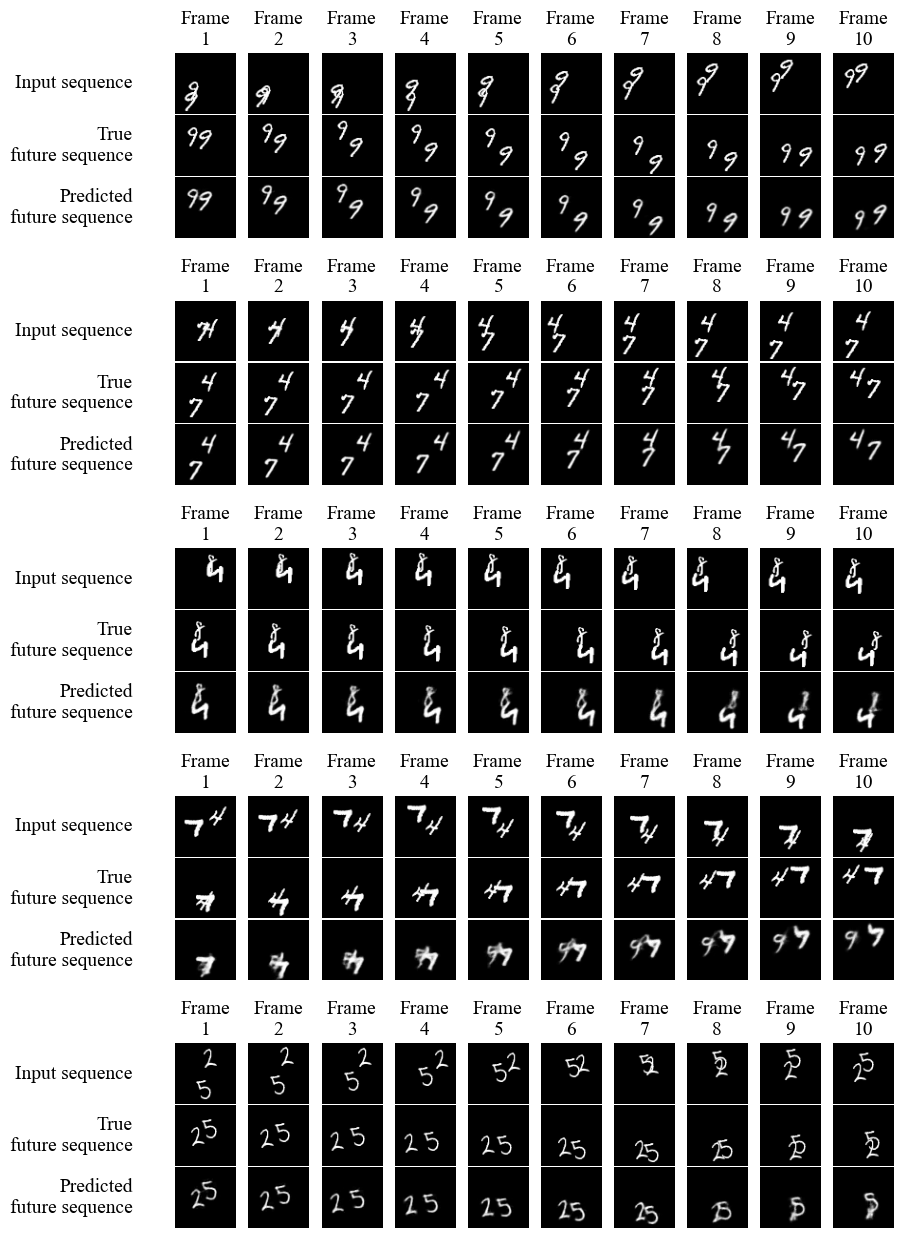}
    \caption{Predictions of the ST-GRU-based PredRNN-V2 on five Moving MNIST test sequences, showing the input sequence, the ground-truth future sequence and the predicted future sequence. Both global and local similarity to the ground truth decline for frames further into the future.}
    \label{fig:results_st-gru_modeloutput}
\end{figure}

\noindent Visual inspection of the generated frames confirms that the model predicts earlier frames (i.e., frames closer to the input frames) more accurately (Figure~\ref{fig:results_st-gru_modeloutput}). Local image features degrade gradually in later frames, especially when the trajectories of the two digits overlap.

\subsection{Quantitative Model Comparison}
Table~\ref{tab:results_modelcomparison} summarises the performance of all models in terms of MSE, SSIM, LPIPS and PSNR, averaged over the ten predicted frames. On MSE and SSIM, the ST-GRU-based PredRNN-V2 performs on par with the ST-LSTM-based PredRNN (MSE 55.1 vs.\ 56.8; SSIM 0.883 vs.\ 0.867). It clearly outperforms ConvLSTM on all metrics, and ConvGRU on SSIM and LPIPS, while its MSE is only slightly lower than that of ConvGRU. The ST-LSTM-based PredRNN-V2 and the ST-GRU-based PredRNN of \citet{dong2025-stgru} achieve better MSE and SSIM than our model. On LPIPS, the ST-GRU-based PredRNN-V2 \emph{does} outperform the ST-LSTM-based PredRNN-V2 (0.065 vs.\ 0.071), and is surpassed only by the ST-GRU-based PredRNN (0.057). On PSNR, which we computed only for the two PredRNN-V2 variants, the ST-LSTM-based model scores higher (20.05 vs.\ 19.53\,dB).

\begin{table}[htb]
\centering
\begin{tabular}{lcccc}
\toprule
Model & MSE ($\downarrow$) & SSIM ($\uparrow$) &  LPIPS ($\downarrow$) & PSNR ($\uparrow$)  \\
\midrule
ConvLSTM (RNN) \citep{shi2015-convlstm} & 103.3 & 0.707 & 0.156 & -- \\
ST-LSTM (PredRNN) \citep{wang2017-predrnn} & 56.8 & 0.867 & 0.107 & -- \\
ST-LSTM (PredRNN-V2) \citep{wang2021-predrnnv2} & \textbf{48.4} & 0.891 & 0.071 & \textbf{20.05} \\
\midrule
ConvGRU (RNN) \citep{ballas2015-convgru} & 57.8 & 0.839 & 0.091 & -- \\
ST-GRU (PredRNN) \citep{dong2025-stgru} & 53.4 & \textbf{0.893} & \textbf{0.057} & -- \\
\textcolor{myteal}{ST-GRU (PredRNN-V2), ours} & \textcolor{myteal}{55.1} & \textcolor{myteal}{0.883} & \textcolor{myteal}{0.065} & \textcolor{myteal}{19.53} \\
\bottomrule
\end{tabular}
\caption{Model performance on the Moving MNIST test set. All metrics are averaged over ten predicted frames. The best result for each metric is shown in bold, and our model in \textcolor{myteal}{teal}. Results for the LSTM-based baselines are taken from \citet{wang2021-predrnnv2} and for the GRU-based baselines from \citet{dong2025-stgru}. PSNR (in dB) was computed only for the two PredRNN-V2 variants.}
\label{tab:results_modelcomparison}
\end{table}

\begin{figure}[htb]
    \centering
    \includegraphics[width=\textwidth]{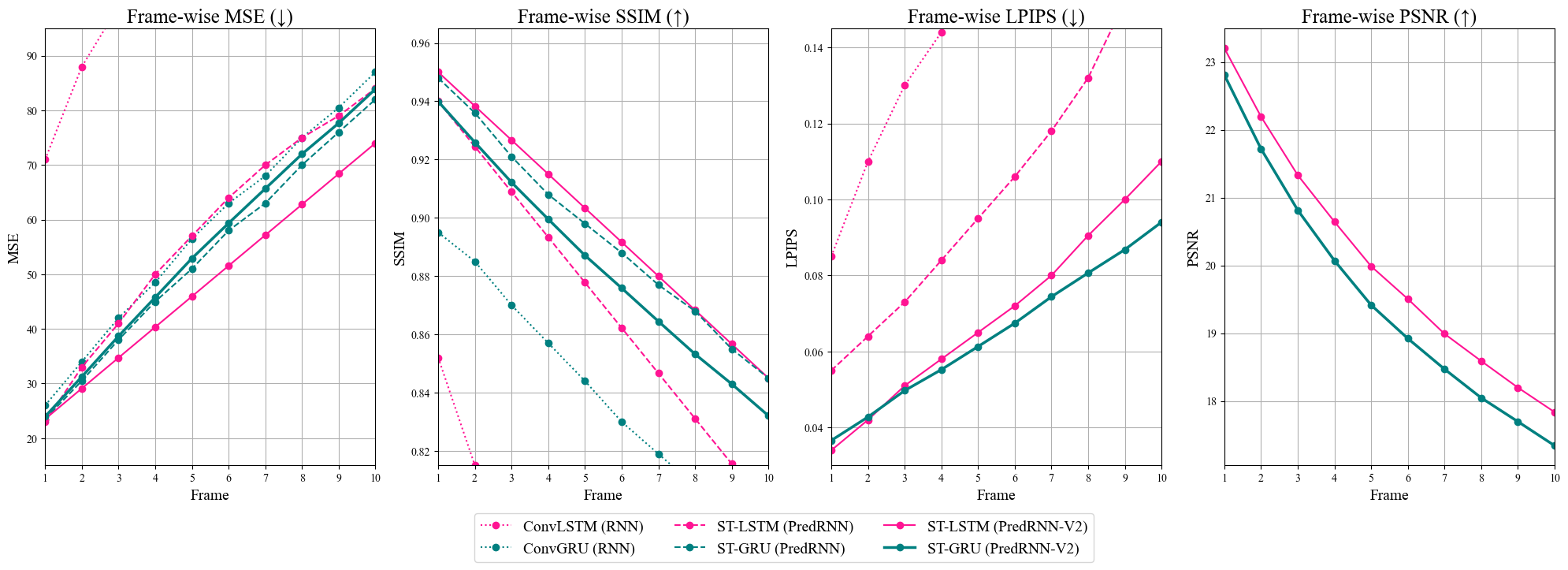}
    \caption{Frame-wise MSE, SSIM, LPIPS and PSNR on the Moving MNIST test set. All models were trained and evaluated following the protocol described in Section~\ref{experimentalsetup}. Results for the LSTM-based models are taken from \citet{wang2021-predrnnv2}, and those for the GRU-based models from \citet{dong2025-stgru}. PSNR for the ST-LSTM-based PredRNN-V2 was computed with the official trained model \citep{github-predrnn}. All results for the ST-GRU-based PredRNN-V2 are from our own evaluation.}
    \label{fig:results_modelcomparison}
\end{figure}

\noindent The frame-wise results in Figure~\ref{fig:results_modelcomparison} agree with these averages. In MSE and SSIM, the ST-GRU-based PredRNN-V2 closely tracks the ST-LSTM-based PredRNN across all ten frames. Frame-wise LPIPS is only available for the LSTM-based baselines; compared with these, our model achieves the lowest LPIPS, and its advantage over the ST-LSTM-based PredRNN-V2 grows for later frames. In PSNR, the ST-LSTM-based PredRNN-V2 is consistently higher, in line with Table~\ref{tab:results_modelcomparison}.

\section{Discussion}
Replacing the ST-LSTM with our ST-GRU in PredRNN-V2 yields a model that performs well across metrics and frames, but that is generally outperformed by the original ST-LSTM-based PredRNN-V2. We believe this is largely due to how the improvements that define PredRNN-V2 interact with our design. The decoupling loss acts on the two separate memory states $\mathcal{C}_t^l$ and $\mathcal{M}_t^l$. Because the ST-GRU does not use $\mathcal{C}_t^l$, this loss cannot be applied in its original form, and we trained the ST-GRU-based model without it. As our experiments did not involve action-conditioned video prediction, the ST-GRU-based PredRNN-V2 therefore benefits mainly from reverse scheduled sampling, whereas the ST-LSTM-based PredRNN-V2 benefits from both reverse scheduled sampling and the decoupling loss. Our results suggest that the simplified cell does not compensate for the loss of this mechanism. Adapting the decoupling loss to act between $\mathcal{H}_t^l$ and $\mathcal{M}_t^l$ is therefore a natural next step.

Interestingly, our model achieves a lower LPIPS than the ST-LSTM-based PredRNN-V2, especially for later frames. LPIPS measures distance in the feature space of a pretrained deep network and is designed to reflect human judgements of perceptual similarity \citep{zhang2018-lpips}. This suggests that the ST-GRU-based model preserves perceptually relevant structure better, even when its pixel-wise error is higher.

The concurrent ST-GRU-based PredRNN of \citet{dong2025-stgru} outperforms our model on all reported metrics, even though it does not use the PredRNN-V2 improvements. This indicates that the specific design of the gated cell matters at least as much as these training improvements. Nevertheless, our ST-GRU-based PredRNN-V2 performs on par with the ST-LSTM-based PredRNN while using a simpler building block. Comparing the two ST-GRU designs under identical conditions, and combining the stronger cell with the PredRNN-V2 training scheme, are promising directions for future work.

\paragraph{Limitations.} We evaluated our model only on Moving MNIST, a synthetic benchmark; experiments on more complex, real-world datasets are needed to establish how well the ST-GRU generalises. We report results from a single training run. Results for the baselines are taken from the literature rather than re-trained under identical conditions. Finally, the ST-GRU reduces the parameter count by only about 5\%, and we did not measure training time, inference time or memory use, so its practical computational benefits remain to be quantified.

\section{Conclusion}
We presented the ST-GRU-based PredRNN-V2, a reformulation of PredRNN-V2 \citep{wang2021-predrnnv2} that replaces its ST-LSTM building blocks with a newly designed ST-GRU cell. The ST-GRU keeps the decoupled spatiotemporal memory flow of the ST-LSTM, but uses fewer gates, one fewer memory state and about 5\% fewer parameters. On Moving MNIST, the model successfully captures the spatiotemporal dynamics of ten-frame future sequences, with accuracy decreasing for later frames, as expected. It performs on par with the ST-LSTM-based PredRNN \citep{wang2017-predrnn} and achieves a lower LPIPS than the ST-LSTM-based PredRNN-V2, although it remains behind PredRNN-V2 on pixel-level metrics. We expect that adapting PredRNN-V2's decoupling loss to the ST-GRU will close part of this gap.

\bibliographystyle{apalike}
\bibliography{ref}

\newpage
\appendix

\section{Model Architectures}
\label{app:ma}

\subsection{Mathematical Comparison}
\label{app:ma_mathmaticalomparison}

\begin{figure}[H]
    \centering
    \small
    \begin{subfigure}[b]{0.45\textwidth}
        \centering
        \begin{align*}
            g_t &= \tanh (W_{xg}*\mathcal{X}_t+W_{hg}*\textcolor{H_blue}{\mathcal{H}_{t-1}^l}) \\
            i_t&=\sigma (W_{xi}*\mathcal{X}_t+W_{hi}*\textcolor{H_blue}{\mathcal{H}_{t-1}^l} ) \\
            f_t&=\sigma (W_{xf}*\mathcal{X}_t+W_{hf}*\textcolor{H_blue}{\mathcal{H}_{t-1}^l} ) \\
            \textcolor{C_green}{\mathcal{C}_t^l}&=f_t\odot \textcolor{C_green}{\mathcal{C}_{t-1}^l}+i_t \odot g_t \\
            \\
            g'_t&= \tanh (W'_{xg}*\mathcal{X}_t + W'_{mg}*\textcolor{M_orange}{\mathcal{M}_{t}^{l-1}}) \\
            i'_t&=\sigma (W'_{xi}*\mathcal{X}_t+W'_{mi}*\textcolor{M_orange}{\mathcal{M}_{t}^{l-1}}) \\
            f'_t&=\sigma (W'_{xf}*\mathcal{X}_t+W'_{mf}*\textcolor{M_orange}{\mathcal{M}_{t}^{l-1}}) \\
            \textcolor{M_orange}{\mathcal{M}_{t}^l}&=f'_t \odot \textcolor{M_orange}{\mathcal{M}_{t}^{l-1}}+i'_t \odot g'_t   \\
            \\
            o_t&=\sigma (W_{xo}*\mathcal{X}_t+W_{ho}*\textcolor{H_blue}{\mathcal{H}_{t-1}^l}\\
            &\quad\quad+W_{co}*\textcolor{C_green}{\mathcal{C}_t^l}+W_{mo}*\textcolor{M_orange}{\mathcal{M}_{t}^l} ) \\
            \textcolor{H_blue}{\mathcal{H}_t^l}&=o_t \odot \tanh(W_{1\times1}*[\textcolor{C_green}{\mathcal{C}_t^l},\textcolor{M_orange}{\mathcal{M}_{t}^l}])
        \end{align*}
        \caption{ST-LSTM}
        \label{fig:ma_mathematicalcomparison_ST-LSTM}
    \end{subfigure}%
    \hfill
    \begin{subfigure}[b]{0.45\textwidth}
        \centering
        \begin{align*}
            z_t&=\sigma(W_{xz}*\mathcal{X}_t+W_{hz}*\textcolor{H_blue}{\mathcal{H}_{t-1}^l}+W_{mz}*\textcolor{M_orange}{\mathcal{M}_{t}^{l-1}}) \\
            r_t&=\sigma(W_{xr}*\mathcal{X}_t+W_{hr}*\textcolor{H_blue}{\mathcal{H}_{t-1}^l}+W_{mr}*\textcolor{M_orange}{\mathcal{M}_{t}^{l-1}}) \\
            \textcolor{H_blue}{\widetilde{\widetilde{\mathcal{H}}}_t^l}&=\tanh(W_{xn}*\mathcal{X}_t+W_{rhn} (r_t \odot (W_{hn}*\textcolor{H_blue}{\mathcal{H}_{t-1}^l}))\\
            &\quad\quad\quad\ +W_{mn}*\textcolor{M_orange}{\mathcal{M}_t^{l-1}}) \\
            \textcolor{H_blue}{\widetilde{\mathcal{H}}_t^l}&=(1-z_t) \odot \textcolor{H_blue}{\widetilde{\widetilde{\mathcal{H}}}_t^l}+z_t\odot \textcolor{H_blue}{\mathcal{H}_{t-1}^l} \\
            \\
            z'_t&=\sigma(W'_{xz}*\mathcal{X}_t+W_{mz}*\textcolor{M_orange}{\mathcal{M}_{t}^{l-1}}) \\
            r'_t&=\sigma(W'_{xr}*\mathcal{X}_t +W_{mr}*\textcolor{M_orange}{\mathcal{M}_{t}^{l-1}}) \\
            \textcolor{M_orange}{\widetilde{\widetilde{\mathcal{M}}}_t^l}&=\tanh(W'_{xn}*\mathcal{X}_t+r'_t\odot(W_{mn}*\textcolor{M_orange}{\mathcal{M}_{t}^{l-1}})) \\
            \textcolor{M_orange}{\mathcal{M}_{t}^l}=\textcolor{M_orange}{\widetilde{\mathcal{M}}_{t}^l}&=(1-z'_t) \odot \textcolor{M_orange}{\widetilde{\widetilde{\mathcal{M}}}_t^l} +z'_t \odot \textcolor{M_orange}{\mathcal{M}_{t}^{l-1}} \\
            \\
            o_t&=\sigma(W_{ho}*\textcolor{H_blue}{\widetilde{\mathcal{H}}_t^l}+W_{mo}*\textcolor{M_orange}{\widetilde{\mathcal{M}}_t^l}) \\
            \textcolor{H_blue}{\mathcal{H}_t^l}&=o_t \odot \tanh(W_{1\times1}*[\textcolor{H_blue}{\widetilde{\mathcal{H}}_t^l},\textcolor{M_orange}{\widetilde{\mathcal{M}}_{t}^l} ])
        \end{align*}
        \caption{ST-GRU}
        \label{fig:ma_mathematicalcomparison_ST-GRU}
    \end{subfigure}
    \caption{Side-by-side mathematical comparison of the ST-LSTM and ST-GRU building blocks.}
    \label{fig:ma_mathematicalcomparison}
\end{figure}

\newpage

\subsection{Visual Comparison}
\label{app:ma_visualcomparison}

\begin{figure}[H]
    \centering
    \begin{subfigure}[t]{0.49\linewidth}
        \centering
        \includegraphics[width=0.9\linewidth]{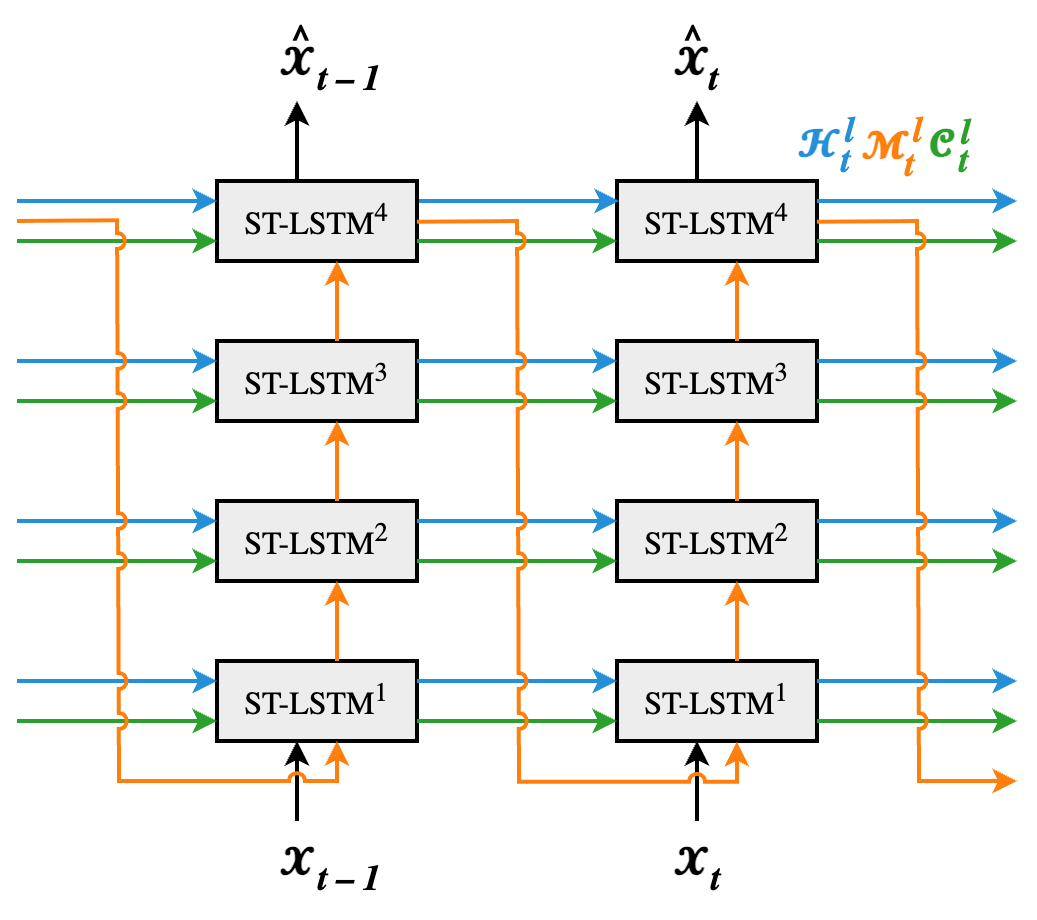}
        \caption{ST-LSTM-based PredRNN(-V2)}
        \label{fig:ma_visualcomparison11_ST-LSTM}
    \end{subfigure}
    \hfill
    \begin{subfigure}[t]{0.49\linewidth}
        \centering
        \includegraphics[width=0.9\linewidth]{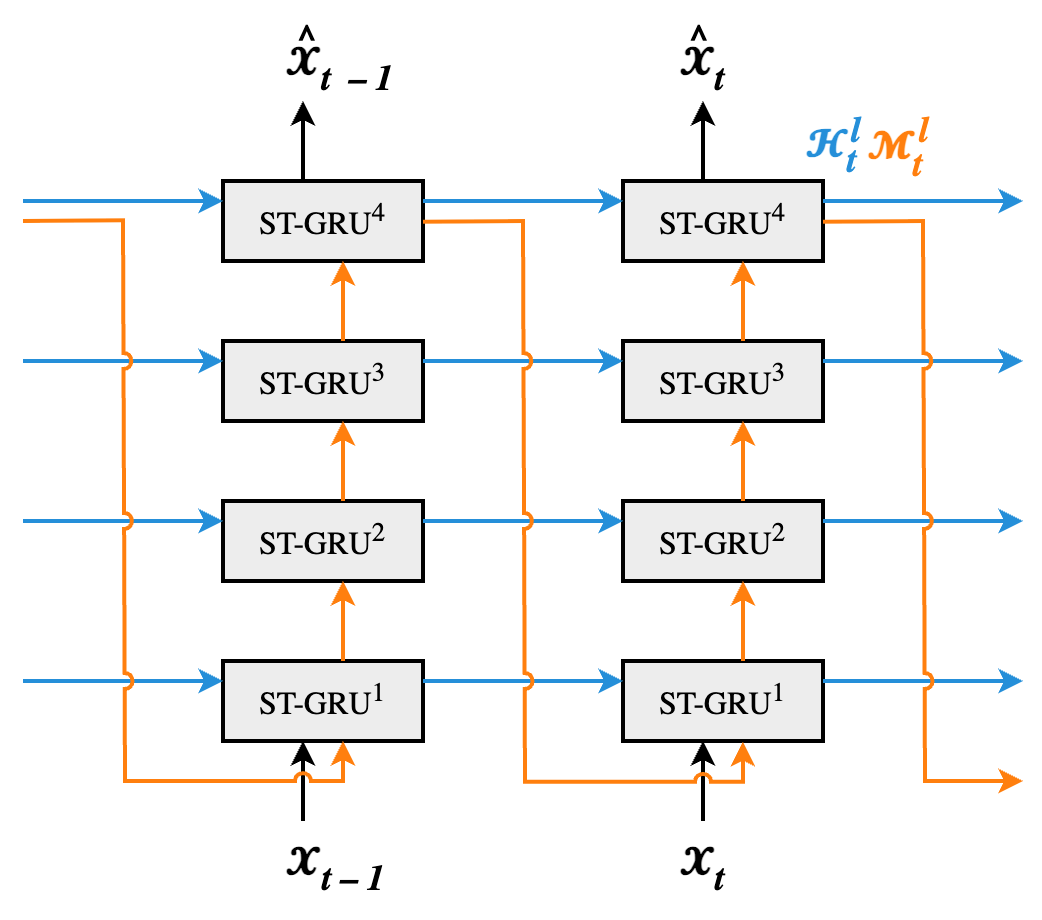}
        \caption{ST-GRU-based PredRNN(-V2)}
        \label{fig:ma_visualcomparison11_ST-GRU}
    \end{subfigure}
    \caption{Side-by-side comparison of the ST-LSTM-based and ST-GRU-based PredRNN(-V2) architectures, with the flows of information shown in different colours. The ST-LSTM-based PredRNN(-V2) has two memory states: the temporal memory \textcolor{C_green}{$\mathcal{C}_t^l$} flows horizontally from the previous time step to the current one ($t$), whereas the spatiotemporal memory \textcolor{M_orange}{$\mathcal{M}_t^l$} zigzags through the layers ($l$) of PredRNN(-V2), capturing short-term dynamics. Both memory states are used to update the hidden state \textcolor{H_blue}{$\mathcal{H}_t^l$}. The ST-GRU-based PredRNN(-V2), in contrast, integrates the horizontal memory flow into the hidden state \textcolor{H_blue}{$\mathcal{H}_t^l$}, while the spatiotemporal memory is still carried by \textcolor{M_orange}{$\mathcal{M}_t^l$}.}
    \label{fig:ma_visualcomparison11}
\end{figure}

\begin{figure}[H]
    \centering
    \begin{subfigure}[t]{0.49\linewidth}
        \centering
        \includegraphics[width=0.9\linewidth]{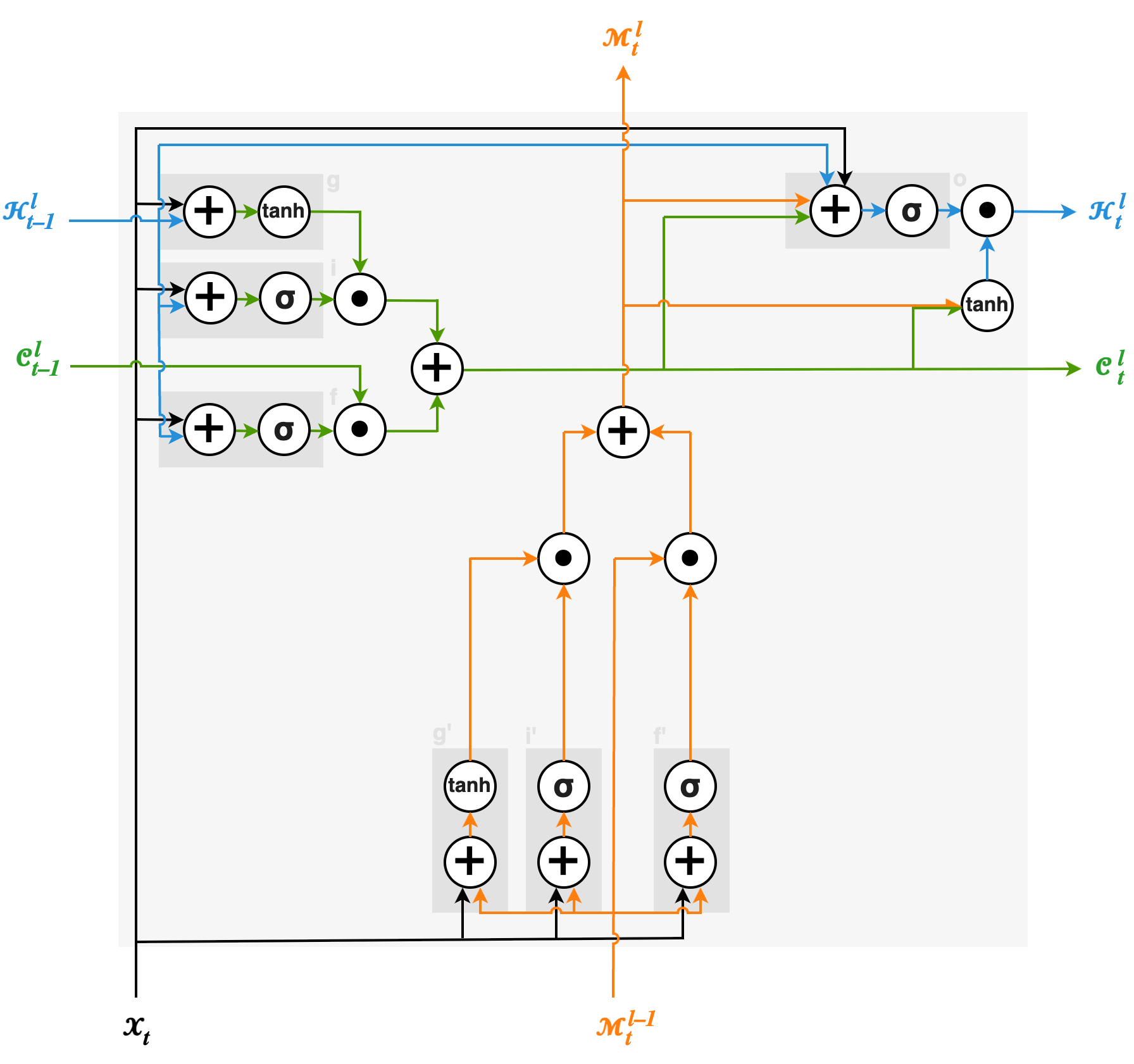}
        \caption{ST-LSTM}
        \label{fig:ma_visualcomparison22_ST-LSTM}
    \end{subfigure}
    \hfill
    \begin{subfigure}[t]{0.49\linewidth}
        \centering
        \includegraphics[width=0.9\linewidth]{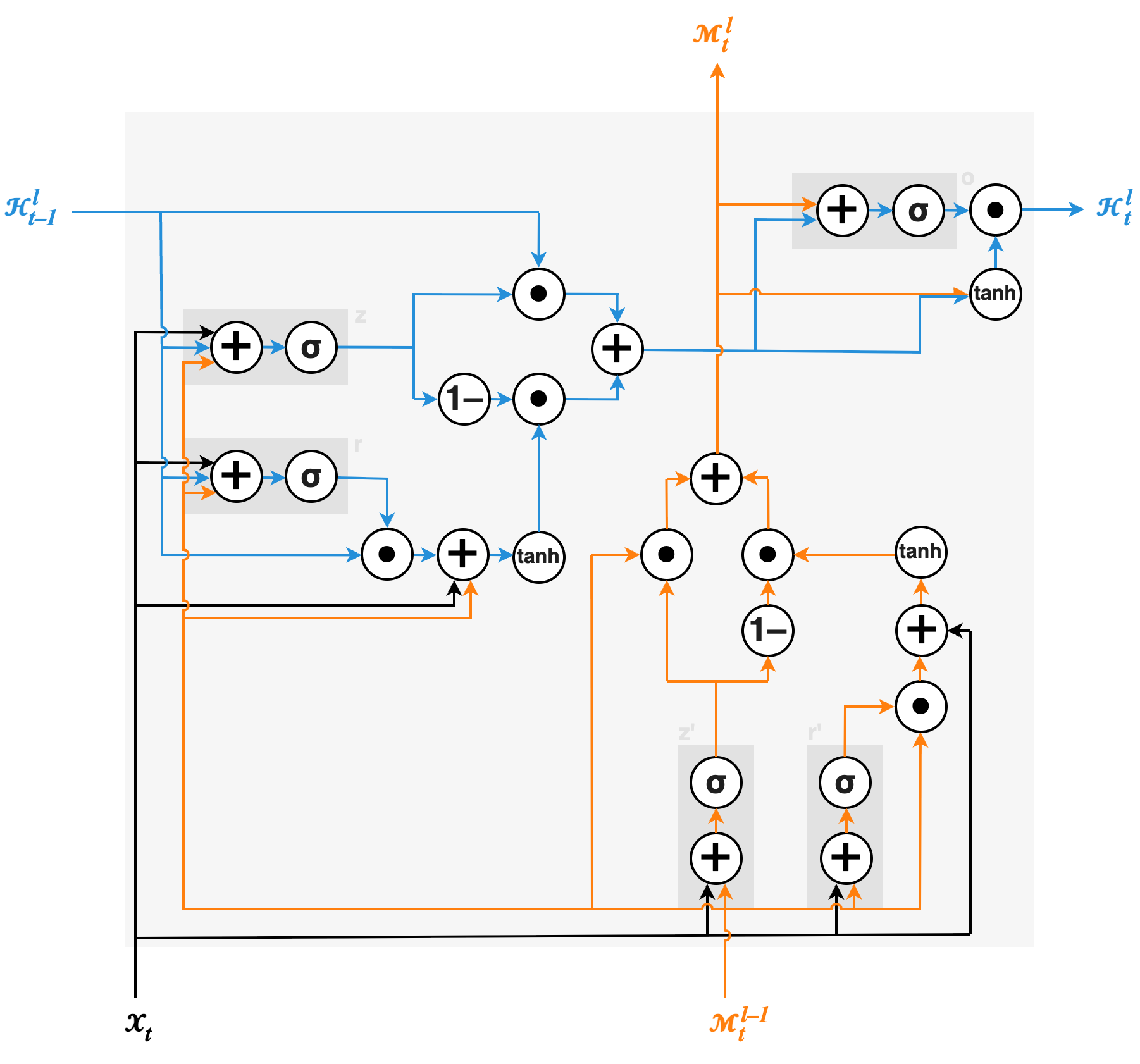}
        \caption{ST-GRU}
        \label{fig:ma_visualcomparison22_ST-GRU}
    \end{subfigure}
    \caption{Side-by-side comparison of the ST-LSTM and ST-GRU building blocks, with the flows of information shown in different colours. The ST-LSTM has two memory states: the temporal memory \textcolor{C_green}{$\mathcal{C}_t^l$} across time steps ($t$), and the spatiotemporal memory \textcolor{M_orange}{$\mathcal{M}_t^l$}, which captures short-term dynamics across layers ($l$). Each memory state is modulated by three gates, $i_t$, $f_t$ and $g_t$ (respectively $i'_t$, $f'_t$ and $g'_t$), and both contribute to the hidden state \textcolor{H_blue}{$\mathcal{H}_t^l$} via the output gate $o_t$. The ST-GRU, in contrast, integrates the horizontal memory flow into the hidden state \textcolor{H_blue}{$\mathcal{H}_t^l$}, while the spatiotemporal memory is still carried by \textcolor{M_orange}{$\mathcal{M}_t^l$}. Each of these states is modulated by two gates, $z_t$ and $r_t$ (respectively $z'_t$ and $r'_t$), and they are combined via the output gate $o_t$. The reduction in both gates and memory states is what makes the ST-GRU simpler than the ST-LSTM.}
    \label{fig:ma_visualcomparison22}
\end{figure}

\end{document}